\documentclass[conference]{IEEEtran}
\IEEEoverridecommandlockouts
\usepackage{amsmath,amssymb,amsfonts}
\usepackage{algorithmic}
\usepackage{graphicx}
\usepackage{textcomp}
\usepackage{xcolor}
\def\BibTeX{{\rm B\kern-.05em{\sc i\kern-.025em b}\kern-.08em
    T\kern-.1667em\lower.7ex\hbox{E}\kern-.125emX}}
\usepackage[pdf]{graphviz}
\usepackage[autosize]{dot2texi}
\usepackage{amsmath}
\usepackage{tikz}
\usepackage{graphicx}
\usepackage{subcaption}
\usepackage{multirow}
\usepackage{cite}
\usepackage[bookmarks=false]{hyperref}
\usepackage[utf8]{inputenc}
\usetikzlibrary{shapes,arrows}

\title{Improving Expressivity of Graph Neural Networks}

\author{\IEEEauthorblockN{Stanisław J. Purgał}
\IEEEauthorblockA{
\textit{University of Innsbruck}\\
Innsbruck, Austria \\
\texttt{stanislaw.purgal@uibk.ac.at}}
}

\begin{document}
\maketitle

\begin{abstract}
We propose a Graph Neural Network with greater expressive power than commonly used GNNs --- not constrained to only differentiate between graphs that Weisfeiler--Lehman test recognizes to be non-isomorphic.
We use a graph attention network with expanding attention window that aggregates information from nodes exponentially far away. We also use partially random initial embeddings, allowing differentiation between nodes that would otherwise look the same. This could cause problem with a traditional dropout mechanism, therefore we use a ``head dropout", randomly ignoring some attention heads rather than some dimensions of the embedding.
\end{abstract}

% keywords can be removed

\section{Introduction}
Recently there has been a great interest in neural network architectures capable of processing graphs \cite{yun2019graph, du2019graph, zhao2018work, ying2019gnnexplainer, xuan2019subgraph}. They are applied for tasks of molecule properties prediction \cite{tox}, premise selection in theorem proving \cite{wang2017premise}, RNA sequence classification \cite{rossi2019ncrna} etc.

Most Graph Neural Networks (GNNs) can recognize graphs only up to Weisfeiler--Lehman isomorphism test (WL-test) \cite{weisfeiler1968reduction, xu2018powerful}, meaning that if the test says the graphs are isomorphic, the networks will process the graphs as if they were exactly the same --- even if they are not.

\begin{figure}
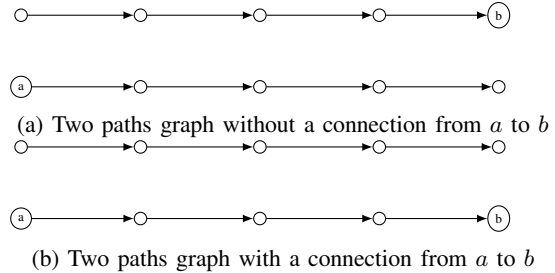

\centering
\begin{subfigure}[a]{0.4\textwidth}
    \begin{dot2tex}[scale=0.5]
    digraph G {
    rankdir=LR;
    0 [label="a"]
    1 [label=""]
    2 [label=""]
    3 [label=""]
    4 [label=""]
    0->1
    1->2
    2->3
    3->4
    
    0x [label=""]
    1x [label=""]
    2x [label=""]
    3x [label=""]
    4x [label="b"]
    0x->1x
    1x->2x
    2x->3x
    3x->4x
    }
    \end{dot2tex}
\caption{Two paths graph without a connection from $a$ to $b$}
\end{subfigure}

\begin{subfigure}[b]{0.4\textwidth}
    \begin{dot2tex}[scale=0.5]
    digraph G {
    rankdir=LR;
    0 [label="a"]
    1 [label=""]
    2 [label=""]
    3 [label=""]
    4 [label="b"]
    0->1
    1->2
    2->3
    3->4
    
    0x [label=""]
    1x [label=""]
    2x [label=""]
    3x [label=""]
    4x [label=""]
    0x->1x
    1x->2x
    2x->3x
    3x->4x
    }
    \end{dot2tex}
\caption{Two paths graph with a connection from $a$ to $b$}
\end{subfigure}
\caption{Graphs consisting of two paths}
    \label{fig:twopaths}
\end{figure}

\begin{figure}
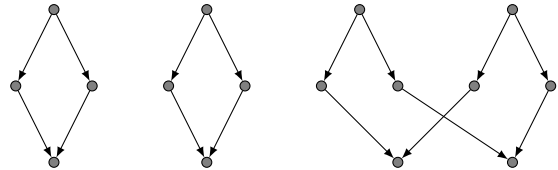

\centering
    \begin{dot2tex}[scale=0.4]
digraph G {
node[style=filled];
0[label=""];
1[label=""];
2[label=""];
3[label=""];
4[label=""];
5[label=""];
6[label=""];
7[label=""]; 0->1; 0->7; 1->2; 3->5; 3->6; 5->4; 6->4; 7->2;

0q[label=""];
1q[label=""];
2q[label=""];
3q[label=""];
4q[label=""];
5q[label=""]; 6q[label=""]; 7q[label=""]; 0q->7q; 1q->6q; 2q->6q; 3q->7q; 4q->2q; 4q->3q; 5q->0q; 5q->1q; }
    \end{dot2tex}
\caption{"Diamond" graphs that common GNNs cannot differentiate}
    \label{fig:diamonds}
\end{figure}

In our work we seek to overcome two types of failure of the WL-test. First is when the difference between graphs is only noticeable when considering long connections (eg. as in fig. \ref{fig:twopaths}). 
Another failure that we correct for is when we need to notice whether two indirect connections lead to one and the same node or to two similar nodes (as in fig. \ref{fig:diamonds}). 

The first failure is addressed in our proposed model by aggregating nodes with an exponentially expanding window. This way we allow the network to notice a connection of exponential length.
This operation could be seen as an attempt to imitate operations done in usual convolutions, such as pooling done in computer vision, which also aggregates information from exponentially far away, although in a more structured way. Another operation we can be said to imitate is an expanding dilated convolution used in WaveNet \cite{oord2016wavenet}, which again aggregates information from far away.
Both those approaches use intrinsic structure of the data to aggregate more information layer by layer rather than trying to process larger and larger sets. Unfortunately, in general, there is no such structure in graphs.

The second problem of the WL-test is solved by introducing a random identifier for every node present in the graph. This preserves invariance under node permutation while allowing the network to differentiate between nodes even if they all look the same --- thus allowing graph attention to be used even when no labels are present.

\section{Preliminaries}
We assume the reader to be familiar with self-attention mechanism \cite{vaswani2017attention} and its use in graph attention networks \cite{velivckovic2017graph}.

Graphs considered in this work are directed, with labelled nodes and edges. We allow all labels in a graph to be equal. Where we consider symmetric graphs, we model it with directed graphs where for every edge there exists a symmetric edge in the other direction. We do not consider multi-edges, though technically our model allows for edges with multiple labels.

When presenting formulas for calculations done in our model we mark parts with learnable parameters with subscript $\phi$. We use $||$ to mark concatenation and $\odot$ to mark point-wise multiplication (or Hadamard product).

\section{Proposed model}
Our proposed model modifies standard graph attention \cite{velivckovic2017graph} in three ways:
\begin{itemize}
\item random initial node embeddings --- to facilitate attention mechanism recognizing different nodes, we add (by concatenation) a random vector to initial embedding of every node, with different random values every time the embeddings are evaluated.
\item expanding attention window --- we use multi-headed attention \cite{vaswani2017attention}, with separate heads for different edge categories. Some of attention heads only see neighbours (as is standard), but some see exponentially expanding neighbourhoods (nodes in distance 2, 4 and so on).
\end{itemize}

\subsection{Partially random initial node embeddings}
In our model (expGNN), initial embedding of a node is composed of two concatenated components of same length. One is a learnable embedding of a node label, the other a \textit{random node identifier}, a random vector composed of $1$s and $0$s (each possible with probability $\frac{1}{2}$).
\begin{align*}
\textbf{n}_{i \phi}^0 &=  \textsc{Embed}_{\phi} (\textsc{Label}(n_i)) \ || \\
&\ \ \ \ \textsc{RandomSequence}(\{ 0: \frac{1}{2}, 1: \frac{1}{2}\}))
\end{align*}
This identifier is different every time an embedding is being calculated, but stays the same withing one graph instance. This means that it is possible to differentiate between nodes, even if their label and neighbourhoods are the same.

\subsection{Expanding attention window}
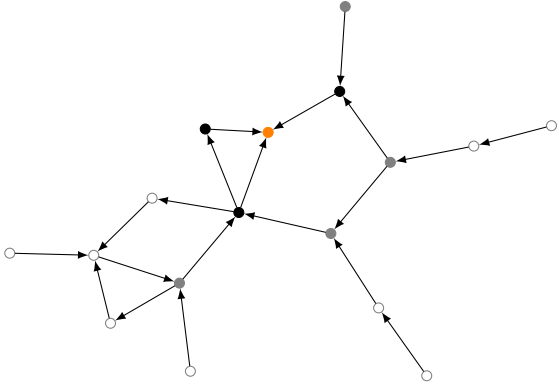
\begin{dot2tex}[neato, scale=0.4]
digraph G{
node [label="",shape=circle, style=filled,color=gray];
0 [color=orange]
a1 [color=black]
a2 [color=black]
a3 [color=black]
c1 [style=""]
c2 [style=""]
c3 [style=""]
c4 [style=""]
d1 [style=""]
d2 [style=""]
d3 [style=""]
d4 [style=""]
d5 [style=""]
a1 -> 0
a2 -> 0
a3 -> 0
a2 -> a3
b1 -> a1
b2 -> a2
b3 -> a2
b4 -> b2
b3 -> d2
a2 -> d4
b4 -> a1
c1 -> b2
c2 -> b3
c3 -> b3
c4 -> b4
d1 -> c2
d2 -> c2
d3 -> c1
d4 -> c2
d5 -> c4
}
\end{dot2tex}

To facilitate propagation of information within a graph (faster than one edge per one layer), we propose an \textit{expanding attention window}. In each layer this window expands exponentially, aggregating information from nodes further away. So, in layer $n$ we aggregate nodes that are within distance $2^n$.

\subsection{Multiple attention filters}
Since it is not clear that this expanding window would be helpful for every task, we use different windows for different attention heads, with some aggregating only neighbours, some using this expanding window, and some aggregating from all nodes in the graph. Since we want information to spread both ways, not only in the direction of edges, we also use different heads where edges go in opposite direction.

All attention head types used in our model are:
\begin{itemize}
    \item Neighbouring nodes (different edge types separately)
    \item Reversed neighbouring nodes (all edge types together)
    \item Expanding window (all edge types together)
    \item Reversed expanding window (all edge types together)
    \item All nodes in the graph
\end{itemize}

When working with an adjacency matrix, expanding the window can be done quite efficiently, by calculating a new adjacency matrix:
\begin{align*}
    A_{n+1} = \min(1, A_n \cdot A_n + A_n)
\end{align*}
Of course, when working with more optimized graph representations for sparse graphs, this operation is very costly, as it makes the graph much denser.

\subsection{Single layer architecture}
For a single layer in our model we use residual connection \cite{he2016deep}, similar to that used in Transformer \cite{vaswani2017attention}, but also utilizing layer normalization \cite{ba2016layer}.

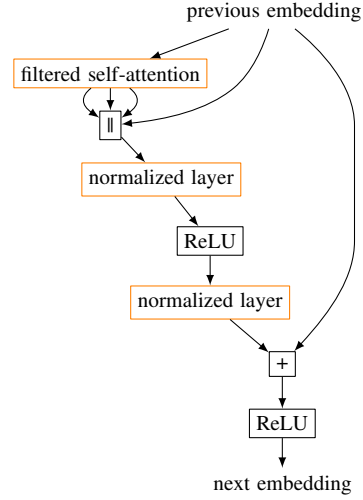
\begin{dot2tex}[scale=0.8]
digraph G {
ranksep=0.2
node [shape=box]
prev [label="previous embedding", color=white]
att [label="filtered self-attention", color=orange]
concat1 [label="‖"]
relu1 [label="ReLU"]
relu2 [label="ReLU"]
nl1 [label="normalized layer", color=orange]
nl2 [label="normalized layer", color=orange]
sum [label="+"]
next [label="next embedding", color=white]
prev -> att
att -> concat1
att -> concat1
att -> concat1
prev -> concat1
concat1 -> nl1
nl1 -> relu1
relu1 -> nl2
prev -> sum
nl2 -> sum
sum -> relu2
relu2 -> next
}
\end{dot2tex}

\begin{align*}
\textbf{n}_{i \phi}^{n+1} &= \textsc{ReLU} (\textbf{n}^n_{i \phi} + \textsc{FNN}_\phi (\textbf{n}_{i \phi}^n || \\
&\ \ \ \ \textsc{FilteredMultiHead}_\phi^n (\textbf{n}_{i \phi}^n, \textbf{n}_{* \phi}^n, \textbf{n}_{* \phi}^n)))
\\ \\
\textsc{FNN}_\phi (x) &= \textsc{NormalizedLayer}_{\phi} ( \\
&\ \ \ \ \textsc{ReLU}(\textsc{NormalizedLayer}_{\phi}(x)))
\end{align*}    

Multi-headed dot-product attention works as in \cite{vaswani2017attention}, only difference being using different masks for different heads.

\begin{align*}
\textsc{Attention}_{\phi} (Q, K, V, M) &= \alpha V W_{\phi}^V \\
\alpha &= \textsc{maskedSoftmax}(M, \beta) \\
\beta &= \frac{ (Q W_{\phi}^Q) (K W_{\phi}^K)^T } {\sqrt{d_k}} \\
\textsc{maskedSoftmax} (M, x) &= \frac{\exp{x} \odot M}{\sum{\exp{x} \odot M}}
\end{align*}
\begin{align*}
\textsc{FilteredMultiHead}_\phi^n (Q, K, V) &= \textsc{concat}( \\
&\ \ \ \ \textsc{Attention}_{\phi} (Q, K, V, M_1) \\
&\ \ \ \ \vdots \\
&\ \ \ \ \textsc{Attention}_{\phi} (Q, K, V, M_h))
\end{align*}

Normalized layer is defined in \cite{ba2016layer} as:
\begin{align*}
\textsc{NormalizedLayer}_\phi (x) &= \frac{g_\phi}{\sigma} \odot (a - \mu) + b_\phi  \\
a &= x A_\phi \\
\mu &= \frac{1}{H} \sum_{i=1}^H a_i \\
\sigma &= \sqrt{\frac{1}{H} \sum_{i=1}^H (a_i - \mu)^2}
\end{align*}

\subsection{Final aggregation for graph classification}
The node embeddings resulting from a few layers described above (in our experiments 3) are aggregated from all nodes in the graph using simple \textit{maximum}.
The resulting graph embedding is fed to a two-layer feed-forward network.

\subsection{Head dropout}
The standard dropout \cite{srivastava2014dropout} mechanism may conflict with the random initial embeddings. The network is supposed to rely on random distribution of vector representations of the nodes in the graph.
Using dropout changes this distribution, making it different during training and during evaluation. This could (and a few times did during the experiments) lead to a situation where loss goes down while the accuracy remains poor.

To counteract this problem, and to force learning of different useful properties, we use a "head dropout". Instead of removing some parts of vectors, we randomly ignore certain attention heads.
During training, each type of attention window (immediate neighbours, expanding, reversed etc.) is ignored with some probability (in our experiments $0.1$).

\section{Experiments}
We test ability of our model to recognize properties that theoretically require overcoming the limitation of WL-test. To do that, we generate artificial datasets, with graph labels determined by the tested property. To better validate generalizing ability, we use more than one evaluation set, with a few different methods of generating random graphs (but using the same property for labels).

For training datasets we use uniform random graphs, where every edge exists with the same probability. This probability is chosen to be such that about half of the generated graphs have the property being tested.

Size of training datasets is $10^6$ (one million), and sizes of random testing datasets are all $10^4$ (ten thousand).
The synthetic datasets used are available online\footnote{\url{http://cl-informatik.uibk.ac.at/cek/ijcnn2020/}} in a format compatible with \cite{datasets}.

\subsection{Presence of a cycle in a symmetric graph}
\label{exp:cycle}
We generate symmetric graphs with 32 nodes, and classify them by checking whether there is a cycle in the graph.

Evaluation sets include:
\begin{itemize}
    \item more random graphs from the same distribution as the training set
    \item uniform random graphs with 64 nodes (with lower edge-existence probability) and with 16 nodes (with higher edge-existence probability)
    \item random trees
    \item random trees with one additional edge (creating a cycle)
    \item line graphs of length between 3 and 64
    \item cycles of length between 3 and 64
\end{itemize}
Random trees are generated by adding nodes one by one, attaching each one to a random already existing node. Half of such generated trees also receive one additional edge between a random pair of not connected nodes.

\subsection{Presence of a clique 4}
\label{exp:clique4}
For training, again, we use random uniform graphs with 16 nodes. Evaluation sets include also bigger (and sparser) graphs than those used in training.

In each set the class on a graph depends on presence of a clique 4 (a subset of 4 nodes where is each node is connected to every other node).

\subsection{Categorizing circulant skip links}
\begin{figure}
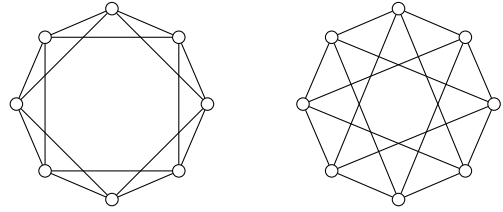

\centering
    \begin{dot2tex}[scale=0.5,neato]
    graph G {
node[shape=circle,label=""]
0 [pos="0,1!"]
1 [pos="0.7,0.7!"]
2 [pos="1,0!"]
3 [pos="0.7,-0.7!"]
4 [pos="0,-1!"]
5 [pos="-0.7,-0.7!"]
6 [pos="-1,0!"]
7 [pos="-0.7,0.7!"]
0 -- 1;
1 -- 2;
2 -- 3;
3 -- 4;
4 -- 5;
5 -- 6;
6 -- 7;
7 -- 0;

0 -- 2;
1 -- 3;
2 -- 4;
3 -- 5;
4 -- 6;
5 -- 7;
6 -- 0;
7 -- 1;

a0 [pos="3,1!"]
a1 [pos="3.7,0.7!"]
a2 [pos="4,0!"]
a3 [pos="3.7,-0.7!"]
a4 [pos="3,-1!"]
a5 [pos="2.3,-0.7!"]
a6 [pos="2,0!"]
a7 [pos="2.3,0.7!"]
a0 -- a1;
a1 -- a2;
a2 -- a3;
a3 -- a4;
a4 -- a5;
a5 -- a6;
a6 -- a7;
a7 -- a0;

a0 -- a3;
a1 -- a4;
a2 -- a5;
a3 -- a6;
a4 -- a7;
a5 -- a0;
a6 -- a1;
a7 -- a2;

}
    \end{dot2tex}
\caption{Circulant skip links --- $\mathcal{G}_\text{skip}(8, 2)$ and $\mathcal{G}_\text{skip}(8, 3)$}
\label{fig:gskips}
\end{figure}

We test our network on the dataset the most difficult dataset used in \cite{murphy2019relational, dasoulas2019coloring}. This dataset has 10 categories, with only 1 graph each.
Each graph is a $\mathcal{G}_\text{skip}(41, R)$ with $R$ being one of $\{ 2,3,4,5,6,9,11,12,13,16 \}$. A graph  $\mathcal{G}_\text{skip}(N, R)$ contains $N$ nodes $\{1, ..., N\}$, such that a pair of nodes $(a, b)$
is connected if (and only if) $|a-b| \equiv 1\ \text{or}\ R\ (\text{mod}\ N)$ (see fig. \ref{fig:gskips} for an example).

In their tests \cite{murphy2019relational, dasoulas2019coloring} use 15 randomly permuted instances of each graph (for a total of 150 graphs in the dataset). Since our network is invariant under permutations, that would be pointless here, and we only use 10 graphs. For evaluation however, since our model is non-deterministic, we do use 150 graphs to get a better evaluation of our accuracy.

Since no generalizing beyond the training set in necessary in this test, we do not use dropout here.

\begin{table*}
\caption{Presence of a clique 4 results}
\label{res:clique}
\begin{center}
\begin{tabular}{ r | c | c c c }
   \multirow{2}{*}{model} & \multicolumn{4}{c}{accuracy} \\
  & training & size 16 & size 32 & size 64 \\
  \hline
  GFN & 0.8076 & 0.8142 & 0.5068 & 0.5092 \\
  GCN & 0.9005 & 0.9088 & 0.5118 & 0.4887 \\
  GraphStar & 0.9983 & \textbf{0.9772} & 0.5331 & 0.5092 \\
  \hline
  expGNN & 0.9129 & 0.9108 & 0.7349 & \textbf{0.5662} \\
  expanding window only & 0.5016 & 0.4924 & 0.4955 & 0.4908 \\
  random init only & 0.9293 & 0.9279 & \textbf{0.7401} & 0.5427 \\
  basic graph attention & 0.5016 & 0.4924 & 0.4955 & 0.4908 \\
\end{tabular}
\end{center}
\end{table*}

\begin{table*}
\caption{Presence of a cycle results}
\label{res:cycle}
\begin{center}
\begin{tabular}{ r | c | c c c c c c c }
   & \multicolumn{7}{c}{accuracy} \\
  & training & uniform 32 & uniform 64 & uniform 16 & trees 64 & trees 32 & lines + cycles\\
  \hline
  GFN & 0.9947 & 0.9961 & 0.9099 & 0.8737 & 0.7659 & 0.8141 & 0.0887 \\
  GCN & 0.9995 & \textbf{0.9996} & 0.8914 & 0.8768 & 0.5649 & 0.8755 & 0.2419 \\
  GraphStar & 0.9983 &  0.9729 & \textbf{0.9994} & \textbf{1.0000} & 0.8167 & 0.9005 & 0.1452 \\
  \hline
  expGNN & 0.9990 & 0.9993 & 0.9819 & 0.9999 & \textbf{0.8227} & \textbf{0.9499} & 0.5726 \\
  expanding window only & 0.5294 & 0.5275 & 0.4424 & 0.5106 & 0.4951 & 0.4982 & 0.5000 \\
  random init only & 0.9823 & 0.9836 & 0.9230 & 0.9972 & 0.7950 & 0.9226 & \textbf{0.6129} \\
  basic graph attention & 0.5294 & 0.5275 & 0.4424 & 0.5106 & 0.4951 & 0.4982 & 0.5000 \\
\end{tabular}
\end{center}
\end{table*}

\begin{table}
\caption{Circulant skip links results}
\label{res:csl}
\begin{center}
\begin{tabular}{ r | c c | c | c }
   \multirow{2}{*}{model} & \multicolumn{4}{c}{accuracy} \\
  & mean & std & max & min \\
  \hline
  RP-GIN \cite{murphy2019relational} & 0.376 & 0.129 & 0.533 & 0.100 \\
  16-CLIP \cite{dasoulas2019coloring}  & 0.908 & 0.068 & 0.987 & 0.760 \\
    Ring-GNN \cite{chen2019equivalence} & N/A & 0.157 & 0.800 & 0.100 \\
  \hline
  expGNN & \textbf{0.978} & 0.015 & \textbf{0.993} & \textbf{0.947} \\
  expanding window only & 0.100 & 0.000 & 0.100 & 0.100 \\
  random init only & 0.687 & 0.030 &  0.740 & 0.647 \\
  basic graph attention & 0.100 & 0.000 & 0.100 & 0.100 \\
\end{tabular}
\end{center}
\end{table}

\begin{table}
\begin{center}
\caption{Presence of node of degree 7 results}
\label{res:degree}
\begin{tabular}{ r | c | c c }
   \multirow{2}{*}{model} & \multicolumn{3}{c}{accuracy} \\
  & training & size 16 & size 32 \\
  \hline
  GFN & 1.0000 & \textbf{1.0000} &  \textbf{1.0000} \\
  GCN &  1.0000 &  \textbf{1.0000} & 0.8300 \\
  GraphStar &  1.0000 & \textbf{1.0000} &  \textbf{1.0000} \\
  \hline
  expGNN & 0.9520 & 0.9534 & 0.5903 \\
  expanding window only & 0.5863 & 0.5906 & 0.5385 \\
  random init only & 0.9862 & 0.9873 & 0.6402 \\
  basic graph attention & 0.5863 & 0.5906 & 0.5385 \\
\end{tabular}
\end{center}
\end{table}

\begin{table*}
\begin{center}
\caption{Presence of a path results} 
\label{res:path}
\begin{tabular}{ r | c | c c c c }
   \multirow{2}{*}{model} & \multicolumn{5}{c}{accuracy} \\
  & training & size 16 & size 32 & size 64 & paths\\
  \hline
  GFN & 0.8358 & 0.8453 & 0.8276 & 0.6775 & 0.5483 \\
  GCN &  0.9706 & 0.7057 & 0.9696 & 0.6810 & 0.5161 \\
  GraphStar & 0.9979 & 1.0000 & 0.9975 & 0.9925 & 0.5967 \\
  \hline
  expGNN & 1.0000 & \textbf{1.0000} & \textbf{1.0000} & \textbf{1.0000} & 0.8371 \\
  expanding window only & 1.0000 & \textbf{1.0000} & \textbf{1.0000} & 0.9999 & \textbf{0.8629} \\
  random init only & 0.9903 & 0.9984 & 0.9889 & 0.9853 & 0.5645 \\
  basic graph attention & 0.9881 & 0.9977 & 0.9866 & 0.9859 & 0.5806 \\
\end{tabular}
\end{center}
\end{table*}

\begin{table}
\caption{Chemical datasets results}
\label{res:chem}
\begin{center}
\begin{tabular}{ r | c c c }
   \multirow{2}{*}{model} & \multicolumn{3}{c}{accuracy} \\
  & SN12C & MOLT-4 & Yeast \\
  \hline
  GFN & 0.9639 & 0.9374 & \textbf{0.8899} \\
  GCN & 0.9592 & 0.9336 & 0.8871 \\
  GraphStar & 0.9640 & \textbf{0.9394} & 0.8884 \\
  \hline
  expGNN & 0.9633 & 0.9335 & 0.8870 \\
  expanding window only & \textbf{0.9656} & 0.9365 & 0.8875 \\
  random init only & 0.9626 & 0.9347 & 0.8865 \\
  basic graph attention & 0.9648 & 0.9365 & 0.8870 \\
\end{tabular}
\end{center}
\end{table}

\subsection{Presence of a path from one highlighted node to the other}
\label{exp:path}
As earlier, training set consists of uniformly random graphs, now with two nodes being given special labels ($a$ and $b$). The class of a graph depends on the existence of a path from $a$ to $b$. In the training set all graphs have 32 nodes.

As a special testing case we use graphs consisting of two paths. The highlighted nodes can be either on the ends on one path, or on two different paths (shown is figure \ref{fig:twopaths}). Those graphs are very similar, and hard for commonly used GNNs to differentiate between. We use paths of length from 2 to 32.

\subsection{Presence of a node with 7 neighbours}
\label{exp:neis}
In this dataset we simply generate uniform graphs and check whether there is a node with degree 7 or greater. For training we use graphs of size 16, for testing we use also bigger graphs of size 32.

\subsection{Chemical datasets}
\label{exp:chem}
We also test our model on a few chemical datasets from \cite{datasets}. These were published on \cite{chemdatasets29}, collected from the PubChem website\footnote{\url{https://pubchem.ncbi.nlm.nih.gov/}}. Each dataset belongs to a certain type of cancer screen with the outcome active or inactive.

\subsection{Tested models}
For comparison with our model we use several recently published graph neural architectures: Graph Feature Network\cite{gfn}, Graph Convolutional Network (using implementation from the same work \cite{gfn}) and Graph Star Net \cite{graphstar}.

The exception is the experiment with circulant skip list, where those networks mathematically can't differentiate between graphs. There we compare with results reported in other papers that also used this dataset \cite{murphy2019relational, dasoulas2019coloring, chen2019equivalence}. Since \cite{chen2019equivalence} does not report mean of their results, we leave it as ``N/A".

We also test variants of our model with only one of the two modifications, as well as without both (making it a Graph Attention Network \cite{velivckovic2017graph}).

\subsection{Hyperparameters}
In our model we use 3 layers of graph message passing. In every layer each node is encoded in 128 dimensions. In dot-product attention the queries and keys have 32 dimensions. Each type of attention head is used thrice. During training each type has a $0.1$ chance of being ignored. 
For optimization we use Adam optimizer \cite{kingma2014adam} with default $\beta_1 = 0.9$, $\beta_2 = 0.999$, $\epsilon = 1\text{e-}7$ and learning rate $1\text{e-}3$.

\section{Results and discussion}

\subsection{Presence of a clique and a cycle}

These two observed graph properties are on one hand simple, on the other according to \cite{xu2018powerful} cannot really be expressed by usual GNNs.
Somewhat surprisingly, results in tables \ref{res:clique} and \ref{res:cycle} show that GNNs still learn to recognize them with high accuracy given a graph of the same size as those in the training set.
However, changing the size of the graph and the density of edges greatly lowers the accuracy, revealing that the learned property is not actually what we wanted.

Our proposed model seems to be able to generalize the property to graphs of different sizes much better.

\subsection{Categorizing circulant skip links}
The results in table \ref{res:csl} show that our model achieves better accuracy than reported in \cite{murphy2019relational, dasoulas2019coloring, chen2019equivalence}.

We see that categorizing long skip links is impossible in 3 layers when not using the expanding attention window. With it however, even 3 layers are enough.

We note that \cite{chen2019equivalence} also reports 100\% accuracy with Ring-GNN-SVN, a variant of Ring-GNN that is given top eigenvalues of adjacency matrices, allowing for trivial classification.

\subsection{Presence of a node with degree 7 and presence of a path}

The last two graph properties are things that can be trivially learned by some GNNs. For most used models, the most basic property is the degree of a node (trivially extracted, or in GFN \cite{gfn} just given as part of the initial embedding). In our model, learning to extract the degree a node is possible, but much harder (and, because it depends on random initial embeddings, remains not 100\% accurate). Instead, the basic property is detecting a connection.

%High accuracy on the "two paths" testing set demonstrates ability of our model to use the far information propagation.

\subsection{Chemical datasets}

Experiments on chemical datasets (results shown in table \ref{res:chem}) show that even though our model has higher theoretical expressive power, it does not improve accuracy on chemical benchmarks.

\section{Related work}
This work seeks to improve Graph Neural Networks. The core idea of GNNs \cite{scarselli2008graph} is to generate new node embeddings by aggregating embeddings of neighbouring nodes. 
Initially, a recurrent networks would process the nodes until their embeddings converged to some value. Currently, most networks use some constant number of layers that aggregate nodes (as do we).
GraphSAGE \cite{sage} experiments with aggregating embeddings using simple functions like mean and maximum.

Following spectral graph theory Kipf et al. \cite{kipf2016semi} propose a Graph Convolution operator. It can be thought of as a sum aggregation, but with embeddings scaled by an inverse of a square root of a node degree ($\frac{1}{\sqrt{d_n}}$ or $d_n^{-\frac{1}{2}}$), both before and after aggregation. In \cite{gfn} some features are added to the initial node embeddings.

Graph Attention Networks, a type of GNNs that we build on, were introduced in \cite{velivckovic2017graph}. In this network attention mechanism \cite{bahdanau2014neural} is used to aggregate the embeddings. Attention extracts information from a set of vectors (representations of things --- in our case nodes) by first estimating importance of every element of the set and then calculating weighted average (with weights depending on importance). 
The importance calculation can be done using a smaller feedforward neural network that given the context and the element estimates importance of the element in the context,
or (as in \cite{vaswani2017attention}) by calculating a dot-product of some projection of context representation with a projection of the element.

The attention mechanism allows for aggregating information of a \textit{set}, rather than a \textit{sequence} (that is, ignore ordering of elements), which fits exactly what we need in graph processing.

All of the above mentioned networks allow information to travel only one edge per network layer. To allow far information  propagation a Graph Star Net was proposed \cite{graphstar}, where a global state (a few "star" nodes) is updated in every layer. This allows information to propagate globally and to neighbours, but not anything in-between. Thus, this network still suffers from the constraint of WL-test. Our model allows also far-but-not-global propagation, it is however much more computationally costly for large graphs.

\subsection{On Graph Attention without node labels}
The output of an attention mechanism with a multiset of exactly the same elements as input will always be equal to that one element. This is because the output of attention is essentially a weighted average, and if all elements are the same, the output will be the same regardless of what (and how many) the weights are. 

Because of this, in a simple graph attention network, when all the nodes have the same embedding (eg. when no node labels are provided), they will remain the same after however many layers. The network can only differentiate between having any neighbours and having none.

The GraphStar network \cite{graphstar} gets around this problem by using attention across both neighbours and a few global stars. In this way, after one layer the embedding depends on the degree of a node (effectively nodes are given labels based on their degree).

Our proposed model uses random node identifiers, making a situation where all nodes have the same embeddings extremely unlikely (effectively impossible).

\begin{figure}
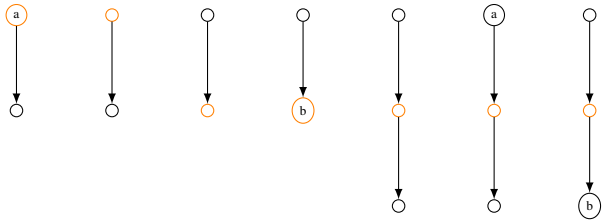

\centering
    \begin{dot2tex}[scale=0.5]
    digraph G {
    0 [label="a", color=orange]
    1 [label=""]
    0 -> 1
    
    0a [label="", color=orange]
    1a [label=""]
    0a -> 1a
    
    0b [label=""]
    1b [label="", color=orange]
    0b -> 1b
    
    0y [label=""];
    1y [label="b", color=orange];
    0y -> 1y

    2x [label=""]
    0x [label="", color=orange]
    1x [label=""]
    2x -> 0x
    0x -> 1x
    
    2aa [label="a"];
    0aa [label="", color=orange];
    1aa [label=""];
    2aa -> 0aa
    0aa -> 1aa
    
    2bb [label=""];
    0bb [label="", color=orange];
    1bb [label="b"];
    2bb -> 0bb
    0bb -> 1bb
    }
    \end{dot2tex}
\caption{All immediate neighbourhoods of a two paths graph (the same regardless of whether $a$ and $b$ are connected)}
\end{figure}

\subsection{Perception limits of GNNs}
Xu et al. \cite{xu2018powerful} describe expressive power of message-passing GNNs as equivalent to Weisfeiler-Lehman isomorphism test \cite{weisfeiler1968reduction}. This means that a node embedding can depend only on the node's subtree structure of certain depth (the depth being equal to the number of layers). The graph classification then depends on the multiset of subtree structures present in the graph. A network capable of distinguishing between all multisets of subtree structures (of certain depth) is referred to in \cite{xu2018powerful} as a \textit{maximally powerful GNN}.
Yet even such networks cannot differentiate between graphs that WL-test deems isomorphic.

The work of \cite{murphy2019relational} seeks to overcome this problem by using by using using a permutation sensitive aggregator and summing over all permutations. To make this computationally feasible, 
they propose $k$-ary Relational Pooling.

A different approach is proposed in \cite{dasoulas2019coloring} where they use one-hot encoded coloring of nodes added in way that allows for differentiating between nodes.

A mechanism similar to our expanding window was described in \cite{chen2017supervised}, where multiple powers of adjacency matrix were used during aggregation (in their experiments they were $A^2$ and $A^4$).
Later in \cite{chen2019equivalence} they use a learnable mechanism to calculate consecutive powers of adjacency matrix, that can in particular learn to express the a property very similar to our exponentially expanding window (what the model in \cite{chen2019equivalence} can express is $\min(A^{2^n}, 1)$, a window containing all nodes that can be reached in \textit{exactly} $2^n$ steps). In this variant, the ``adjacency matrices" don't necessarily contain only 1s and 0s.

Our work seeks to expand the limit of WL-test in two places: for one, by utilizing expanding attention window we effectively increase depth of subtree structures exponentially. Since we also use direct neighbourhood in some attention heads we theoretically don't lose any expressive power.
Another way our model is more expressive is its ability to recognize connection to one and the same node from a connection to two identical nodes. We achieve this by using random initial embeddings.

We should point out that because of use of randomness we lose the property of isomorphic graphs always having the same embedding --- instead we have isomorphic graphs having the same \textit{distribution} of embeddings.

\section{Conclusion}
We present a Graph Neural Network with more expressive power than any model we have seen described. We show its ability to differentiate between graphs that other networks cannot. 
What our models seems to excel at is classifying synthetic datasets of graphs WL-test fails to recognize as non-isomorphic and generalization to previously unseen graph sizes (and edge densities).

Future work includes translating this improvement to accuracy on chemical datasets.

\section*{Acknowledgement}

This research was supported by the ERC starting grant no. 714034 SMART.

\bibliographystyle{IEEEtran}
\bibliography{IEEEabrv,references}

\end{document}